\documentclass[10pt,twocolumn,letterpaper]{article}

\usepackage{cvpr}              
\usepackage{graphicx}

\usepackage[dvipsnames]{xcolor}

\definecolor{cvprblue}{rgb}{0.21,0.49,0.74}
\usepackage[pagebackref,breaklinks,colorlinks,citecolor=cvprblue]{hyperref}

\def\paperID{6983} 
\def\confName{CVPR}
\def\confYear{2024}

\title{GaussianBody: Clothed Human Reconstruction via 3d Gaussian Splatting\\} 

\author{Mengtian Li$^{1,3}$, 
Shengxiang Yao$^{1}$, 
Zhifeng Xie$^{1,3,}$\footnotemark[2], 
Keyu Chen$^{2}$\footnotemark[2]\\
$^{1}$Shanghai University, $^{2}$Tavus Inc.\\
$^{3}$Shanghai Engineering Research Center of Motion Picture Special Effects\\
\tt\small \{mtli,yaosx033,zhifeng\_xie\}@shu.edu.cn, keyu@tavus.dev
}

\begin{document}

\twocolumn[{%
\renewcommand\twocolumn[1][]{#1}%
\maketitle
\begin{center}
    \centering
    \captionsetup{type=figure}
    \includegraphics[width=\textwidth]{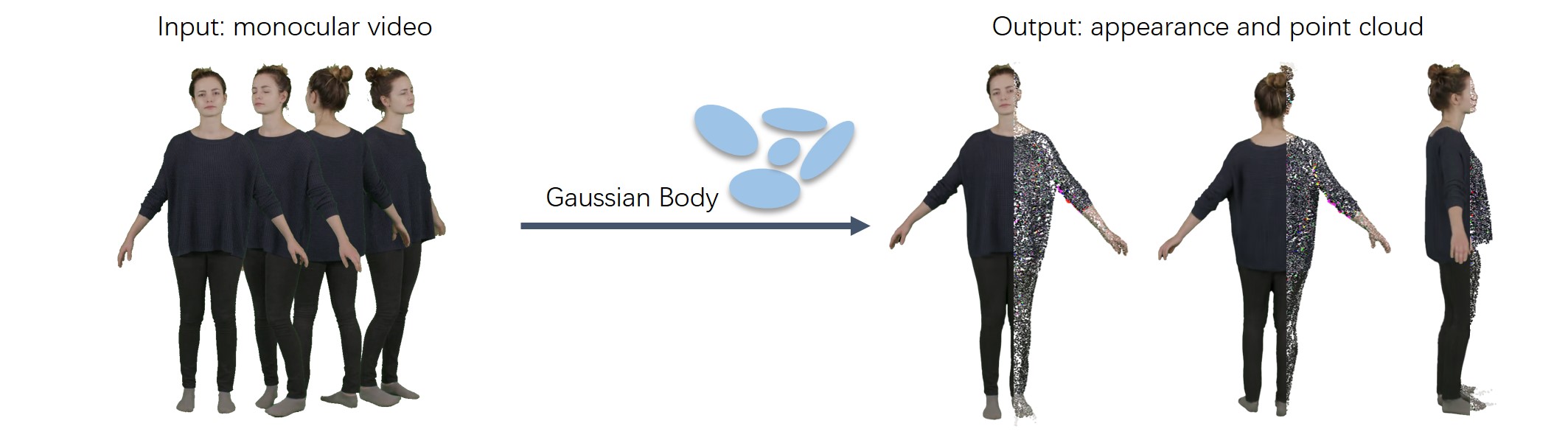}
    \captionof{figure}{GaussianBody takes monocular RGB video as input, reconstructing a clothed human model from 1080$\times$1080 images in around 1 hour on a single 4090 GPU. The resulting human model serves as a tool for simulating human performance in novel views. Furthermore, we offer the point cloud as a mechanism for deformation control.}
\end{center}%
}]
\renewcommand{\thefootnote}{\fnsymbol{footnote}}
\footnotetext[2]{Corresponding author.}

\begin{abstract}
In this work, we propose a novel clothed human reconstruction method called GaussianBody, based on 3D Gaussian Splatting. 
Compared with the costly neural radiance-based models, 3D Gaussian Splatting has recently demonstrated great performance in terms of training time and rendering quality. 
However, applying the static 3D Gaussian Splatting model to the dynamic human reconstruction problem is non-trivial due to complicated non-rigid deformations and rich cloth details.
To address these challenges, our method considers explicit pose-guided deformation to associate dynamic Gaussians across the canonical space and the observation space, introducing a physically-based prior with regularized transformations helps mitigate ambiguity between the two spaces. 
During the training process, we further propose a pose refinement strategy to update the pose regression for compensating the inaccurate initial estimation and a split-with-scale mechanism to enhance the density of regressed point clouds.
The experiments validate that our method can achieve state-of-the-art photorealistic novel-view rendering results with high-quality details for dynamic clothed human bodies, along with explicit geometry reconstruction.
\end{abstract}
\section{Introduction}
\label{sec:Introduction}
Creating high-fidelity clothed human models holds significant applications in virtual reality, telepresence, and movie production. Traditional methods involve either complex capture systems or tedious manual work from 3D artists, making them time-consuming and expensive, thus limiting scalability for novice users. Recently, there has been a growing focus on automatically reconstructing clothed human models from single RGB images or monocular videos.

Mesh-based methods \cite{kolotouros2019learning, kocabas2020vibe, sun2021monocular, feng2021collaborative} are initially introduced to recover human body shapes by regressing on parametric models such as \textit{SCAPE} \cite{anguelov2005scape}, \textit{SMPL} \cite{SMPL:2015}, \textit{SMPL-X} \cite{pavlakos2019expressive}, and \textit{STAR} \cite{osman2020star}. While they can achieve fast and robust reconstruction, the regressed polygon meshes struggle to capture variant geometric details and rich clothing features. The addition of vertex offsets becomes an enhancement solution \cite{ma2020learning, alldieck2018video} in this context. However, its representation ability is still strictly constrained by mesh resolutions and generally fails in loose-cloth cases.

To overcome the limitations of explicit mesh models, implicit methods based on occupancy fields \cite{saito2019pifu, saito2020pifuhd}, signed distance fields (SDF) \cite{xiu2022icon}, and neural radiance fields (NeRFs) \cite{mildenhall2021nerf, peng2021neural, chen2021animatable, jiang2023instantavatar, weng2022humannerf, li2022tava, li2023posevocab, isik2023humanrf} have been developed to learn the clothed human body using volume rendering techniques. These methods are capable of enhancing the reconstruction fidelity and rendering quality of 3D clothed humans, advancing the realistic modeling of geometry and appearance. Despite performance improvements, implicit models still face challenges due to the complex volume rendering process, resulting in long training times and hindering interactive rendering for real-time applications. Most importantly, native implicit approaches lack an efficient deformation scheme to handle complicated body movements in dynamic sequences \cite{weng2022humannerf, li2022tava, li2023posevocab, isik2023humanrf}. 

Therefore, combining explicit geometry primitives with implicit models has become a trending idea in recent works. For instance, point-based NeRFs \cite{xu2022point, yu2023pointbased} propose controlling volume-based representations with point cloud proxy. Unfortunately, estimating an accurate point cloud from multi-view images is practically challenging as well due to the intrinsic difficulties of the multi-view stereo (MVS) problem.

In this work, we address the mentioned issues by incorporating 3D Gaussian Splatting (3D-GS) \cite{kerbl20233d} into the dynamic clothed human reconstruction framework. 3D-GS establishes a differential rendering pipeline to facilitate scene modeling, notably reducing a significant amount of training time. It learns the explicit point-based model while rendering high-quality results with spherical harmonics (SH) representation. The application of 3D-GS to present 4D scenes has demonstrated superior results \cite{yang2023deformable3dgs,yang2023realtime,wu20234dgaussians}, motivating our endeavor to integrate 3D-GS into human body reconstruction. However, learning dynamic clothed body reconstruction is more challenging than other use cases, primarily due to non-rigid deformations of body parts and the need to capture accurate details of the human body and clothing, especially for loose outfits like skirts.

Firstly, we extended the 3D-GS representation to clothed human reconstruction by utilizing an articulated human model for guidance. Specifically, we use forward linear blend skinning (LBS) to deform the Gaussians from the canonical space to each observation space per frame. Secondly, we optimize a physically-based prior for the Gaussians in the observation space to mitigate the risk of overfitting Gaussian parameters. We transform the local rigidity loss \cite{luiten2023dynamic} to regularize over-rotation across the canonical and observation space. Finally, we propose a split-with-scale strategy to enhance point cloud density and a pose refinement approach to address the texture blurring issue.

We evaluate our proposed framework on monocular videos of dynamic clothed humans. By comparing it with baseline approaches and other works, our method achieves superior reconstruction quality in rendering details and geometry recovery, while requiring much less training time (approximately one hour) and almost real-time rendering speed. We also conduct ablation studies to validate the effectiveness of each component in our method.


\begin{figure*}[ht]
    \centering
    \includegraphics[width=1\linewidth]{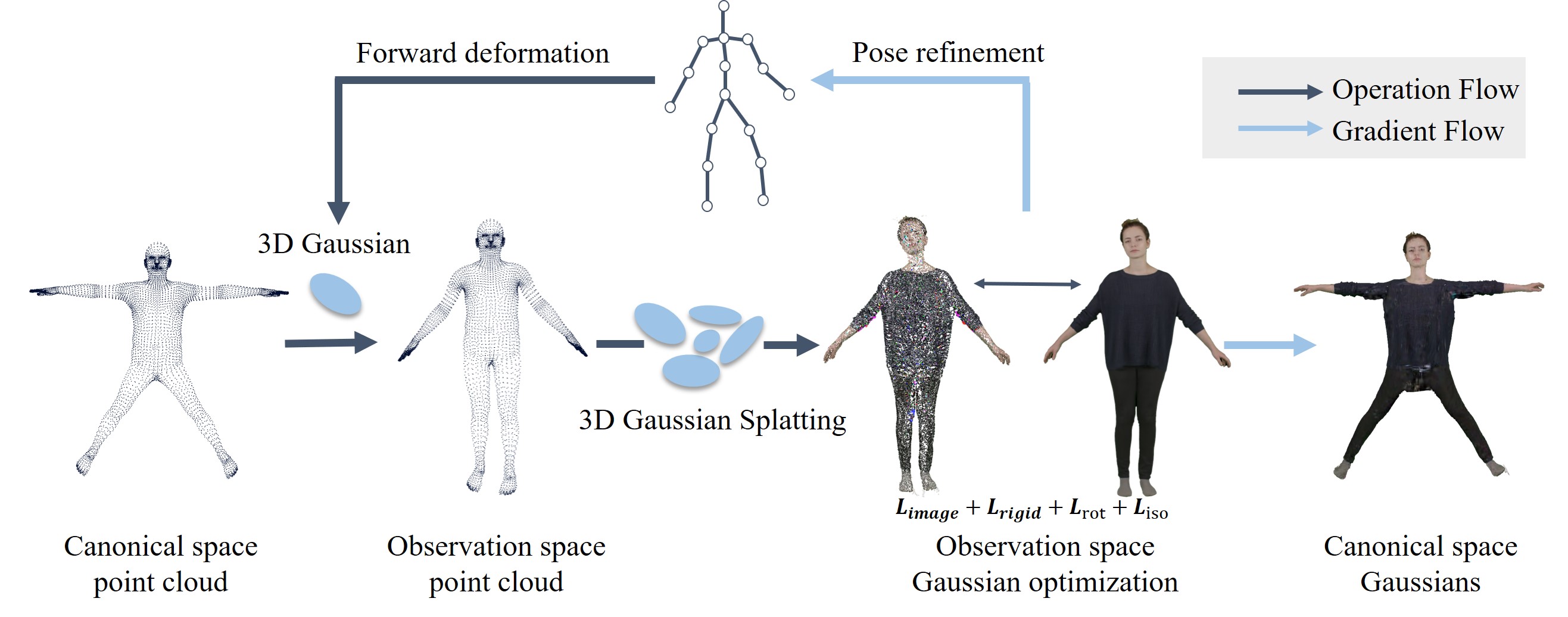}
    \caption{\textbf{Overview of our pipeline.} 
    We initialize the point cloud using SMPL vertices, deforming the position and rotation parameters of Gaussians through SMPL forward linear blend skinning (LBS) to transform them into the observation space. The canonical model is then optimized, taking into account the physically-based prior $\mathcal{L}_{rigid}, \mathcal{L}_{rot}, \mathcal{L}_{iso}$. To address image blurriness, we optimize the pose parameters. The output includes both the point cloud and the appearance of the reconstructed human.
    }
    \label{fig:pipeline}
\end{figure*}

\section{Related Work}
\label{sec:relatedwork}
In this section, we briefly review the related literature with 3D clothed human reconstruction.
\subsection{3D Human Reconstruction}
Reconstructing 3D humans from images or videos is a challenging task. Recent works \cite{alldieck2018detailed,alldieck2018video,ma2020learning} use template mesh models like SMPL \cite{SMPL:2015} to reconstruct 3D humans from monocular videos or single images. However, template mesh models have limitations in capturing intricate clothing details. To address these limitations, neural representations have been introduced \cite{saito2019pifu,saito2020pifuhd,han2023high} for 3D human reconstruction. Implicit representations, like those used in PIFU \cite{saito2019pifu} and its variants, achieve impressive results in handling complex details such as hairstyle and clothing. Some methods, like ICON \cite{xiu2022icon} and ECON \cite{xiu2023econ}, leverage SMPL as a prior to handle extreme poses. However, most of these methods are designed for static scenes and struggle with dynamic scenarios. Other methods \cite{zheng2021pamir,huang2020arch,he2021arch++} use parametric models to handle dynamic scenes and obtain animatable 3D human models.

Recent advancements involve using neural networks for representing dynamic human models. Extensions of NeRF \cite{mildenhall2021nerf} into dynamic scenes \cite{pumarola2021d,park2021nerfies,park2021hypernerf} and methods for animatable 3D human models in multi-view scenarios \cite{isik2023humanrf,lin2023im4d,peng2021animatable,li2022tava,li2023posevocab,weng2022humannerf} or monocular videos \cite{zhao2022human,peng2021neural,chen2021animatable,jiang2023instantavatar} have shown promising results. Signal Distance Function (SDF) is also employed \cite{liao2023high,jiang2022selfrecon,guo2023vid2avatar} to establish a differentiable rendering framework or use NeRF-based volume rendering to estimate the surface. Our method enhances both speed and robustness by incorporating 3D-GS \cite{kerbl20233d}.

\subsection{Accelerating Neural Rendering}
Several methods \cite{hedman2021baking,yu2021plenoctrees,reiser2021kilonerf,chen2023mobilenerf} focus on accelerating rendering speed, primarily using explicit representations or baking methods. However, these approaches are tailored for static scenes. Some works \cite{peng2023representing,wang2022fourier} aim to accelerate rendering in dynamic scenes, but they often require dense input images or additional geometry priors. InstantAvatar \cite{jiang2023instantavatar}, based on instant-NGP \cite{muller2022instant}, combines grid skip rendering and a quick deformation method \cite{chen2023fast} but relies on accurate pose guidance for articulate weighting training. In contrast, 3D-GS \cite{kerbl20233d} offers fast convergence and easy integration into graphics rendering pipelines, providing a point cloud for explicit deformation. Our method extends 3D-GS for human reconstruction, achieving high-quality results and fast rendering.

\section{GaussianBody}
In this section, we first introduce the preliminary method 3D-GS \cite{kerbl20233d} in Section \ref{subsec:pre}. Next, we describe our framework pipeline for 3D-GS-based clothed body reconstruction (Section \ref{subsec:GaussiansBody}). We then discuss the application of a physically-based prior to regularize the 3D Gaussians across the canonical and observation spaces (Section \ref{subsec:PBP}). Finally, we introduce two strategies, split-with-scale and pose refinement, to enhance point cloud density and optimize the SMPL parameters, respectively (Section \ref{subsec:Optimization}).

\subsection{Preliminary}
\label{subsec:pre}
3D-GS \cite{kerbl20233d} is an explicit 3D scene reconstruction method designed for multi-view images. The static model comprises a list of Gaussians with a point cloud at its center. Gaussians are defined by a covariance matrix $\Sigma$ and a center point $X$, representing the mean value of the Gaussian:

\begin{equation}
    G(X) = e^{-\frac{1}{2}X^T\Sigma^{-1}X}.
    \label{eq:Gaussian}
\end{equation}

For differentiable optimization, the covariance matrix $\Sigma$ can be decomposed into a scaling matrix $S$ and a rotation matrix $R$:

\begin{equation}
    \Sigma = RSS^TR^T,
    \label{eq:RandS}
\end{equation}

The gradient flow computation during training is detailed in \cite{kerbl20233d}. To render the scene, the regressed Gaussians can be projected into camera space with the covariance matrix $\Sigma'$:

\begin{equation}
    \Sigma' = JW\Sigma W^TJ^T,
    \label{eq:jacobi}
\end{equation}

Here, $J$ is the Jacobian of the affine approximation of the projective transformation, and $W$ is the world-to-camera matrix. To simplify the expression, the matrices $R$ and $S$ are preserved as rotation parameter $r$ and scaling parameter $s$. After projecting the 3D Gaussians to 2D, the alpha-blending rendering based on point clouds bears a resemblance to the volumetric rendering equation of NeRF \cite{mildenhall2021nerf} in terms of its formulation. During volume rendering, each Gaussian is defined by an opacity $\alpha$ and spherical harmonics coefficients $c$ to represent the color. The volumetric rendering equation for each pixel contributed by Gaussians is given by:

\begin{equation}
    C = \sum_{\substack{i\in \mathbf{N}}}c_i\alpha_i,
    \label{eq:color}
\end{equation}

Collectively, the 3D Gaussians are denoted as $G(x,r,s,\alpha,c)$.

\subsection{Framework}
\label{subsec:GaussiansBody}

In our framework, we decompose the dynamic clothed human modeling problem into the canonical space and the motion space. First, We define the template 3D Gaussians in the canonical space as $G(\bar{x},\bar{r},\bar{s},\bar{\alpha},\bar{c})$. To learn the template 3D Gaussians, we employ pose-guidance deformation fields to transform them into the observation space and render the scene using differentiable rendering. The gradients of the pose transformation are recorded for each time and used for backward optimization in the canonical space. 

Specifically, we utilize the parametric body model SMPL \cite{SMPL:2015} as pose guidance. The articulated SMPL model $M(\beta,\theta)$ is defined with pose parameters $\theta \in R^{3n_k+3}$ and shape parameters $\beta \in R^{10}$. The transformation of each point is calculated with the skinning weight field $w(\beta)$ and the target bone transformation $B(\theta) = \{B_1(\theta), ..., B_n(\theta)\}$. To mitigate computational costs, we adopt the approach from InstantAvatar \cite{jiang2023instantavatar}, which diffuses the skinning weight of the SMPL \cite{SMPL:2015} model vertex into a voxel grid. The weight of each point is then obtained through trilinear interpolation from the grid weighting, denoted as $w(\bar{x}),\beta$. The transformation of the canonical points to deform space via forward linear blend skinning is expressed as:

\begin{equation}
    \mathcal{D}(\bar{x},\theta,\beta) = \sum_{i=1}^{n}w_i(\bar{x},\beta)B_i(\theta).
    \label{eq:deform}
\end{equation}

With the requirements of the 3D-GS\cite{kerbl20233d} initial setting, we initialize the point cloud with the template SMPL\cite{SMPL:2015} model vertex in the canonical pose(as shown in Figure.\ref{fig:pipeline}).
For each frame, we deform the position $\bar{x}$ and rotation $\bar{r}$ of the canonical Gaussians $G(\bar{x},\bar{r},\bar{s},\bar{\alpha},\bar{c})$ with the pose parameter $\theta_t$ of current frame and the global shape parameter $\beta$ to the observation space :
\begin{equation}
\begin{split}
    x &= \mathcal{D}(\bar{x},\theta_t,\beta)\bar{x},\\
    r &= \mathcal{D}(\bar{x},\theta_t,\beta)\bar{r}
\end{split}
    \label{eq:Gaussiandeform}
\end{equation}
where $\mathcal{D}$ is the deformation function defined in Eq.\ref{eq:deform}. 

In this way, we obtain the deformed Gaussians in the observation space. After differentiable rendering and image loss calculation, the gradients will be passed through the inverse of the deformation field $\mathcal{D}$ and optimized for the canonical Gaussians.

\begin{figure}
    \centering
    \includegraphics[width=\linewidth]{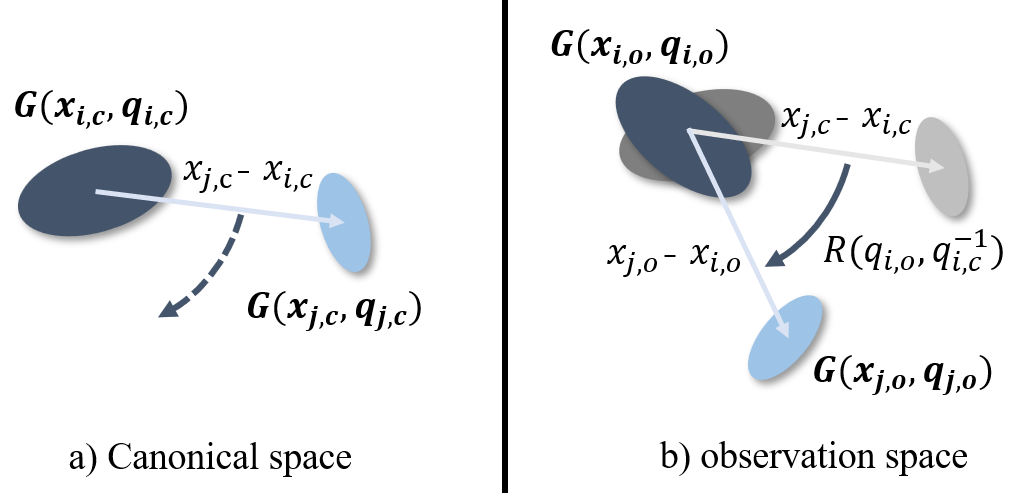}
    \caption{\textbf{Local-rigidity loss.} With the Gaussians $i$ rotating between the two spaces, the neighbour Gaussians $j$ should move to follow the rigid-transform in the coordinate system of Gaussians $i$}
    \label{fig:rigid}
\end{figure}

\subsection{Physically-Based Prior}
\label{subsec:PBP}
Although we define the canonical Gaussians and explicitly deform them to the observation space for differentiable rendering, the optimization is still an ill-posed problem because there could be multiple canonical positions mapped to the same observation position, leading to overfitting in the observation space and visual artifacts in the canonical space.
In the experiment, we also observed that this optimization approach might easily result in the novel view synthesis showcasing numerous Gaussians in incorrect rotations, consequently generating unexpected glitches.
Thus we follow \cite{luiten2023dynamic} to regularize the movement of 3D Gaussians by their local information. Particularly we employ three regularization losses to maintain the local geometry property of the deformed 3D Gaussians, including local-rigidity loss $\mathcal{L}_{rigid}$, local-rotation loss $\mathcal{L}_{rot}$ losses and a local-isometry loss $\mathcal{L}_{iso}$.
Different from \cite{luiten2023dynamic} that attempts to track the Gaussians frame by frame, we regularize the Gaussian transformation from the canonical space to the observation space.

Given the set of Gaussians $j$ with the k-nearest-neighbors of $i$ in canonical space (k=20), the isotropic weighting factor between the nearby Gaussians is calculated as:
\begin{equation}
    w_{i,j} = exp(-\lambda_{w}||x_{j,c}-x_{i,c}||^{2}_{2}),
    \label{eq:local-weight}
\end{equation}
where $||x_{j,c}-x_{i,c}||$ is the distance between the Gasussians $i$ and Gasussians $j$ in canonical space, set $\lambda_{w} = 2000$ that gives a standard deviation.
Such weight ensures that rigidity loss is enforced locally and still allows global non-rigid reconstruction.
The local rigidity loss is defined as:
\begin{equation}
    \mathcal{L}^{rigid}_{i,j} = w_{i,j}||(x_{j,o}-x_{i,o}-R_{i,o}R_{i,c}^{-1}(x_{j,c}-x_{i,c}))||_{2},
    \label{eq:local-rigidity}
\end{equation}
\begin{equation}
    \mathcal{L}_{rigid} = \frac{1}{k|\mathcal{S}|}\sum_{i \in \mathcal{S}}\sum_{j \in knn_{i;k}}\mathcal{L}^{rigid}_{i,j},
    \label{eq:all-rigidity}
\end{equation}
when the Gaussians $i$ transform from canonical space to observation space, the nearby Gaussians $j$ should move in a similar way that follows the rigid-body transform of the coordinate system of the Gaussians $i$ between two spaces.
The visual explanation is shown in Figure.\ref{fig:rigid-loss}.

While the rigid loss ensures that Gaussians $i$ and Gaussians $j$ share the same rotation, the rotation loss could enhance convergence to explicitly enforce identical rotations among neighboring Gaussians in both spaces:
\begin{equation}
    \mathcal{L}_{rot} =  \frac{1}{k|\mathcal{S}|}\sum_{i \in \mathcal{S}}\sum_{j \in knn_{i;k}}w_{i,j}||q_{j,o}q^{-1}_{j,c}-q_{i,o}q^{-1}_{i,c}||_{2},
    \label{eq:local-rot}
\end{equation}
where $q$ is the normalized Quaternion representation of each Gaussian's rotation, the $q_{o}q^{-1}_{c}$ demonstrates the rotation of the Gaussians with the deformation.
We use the same Gaussian pair sets and weighting function as before.

Finally, we use a weaker constraint than $\mathcal{L}^{rigid}$ to make two Gaussians in different space to be the same one, which only enforces the distances between their neighbors:
\begin{equation}
    \mathcal{L}_{iso}=\frac{1}{k|\mathcal{S}|}\sum_{i \in \mathcal{S}}\sum_{j \in knn_{i;k}}w_{i,j}\{||x_{j,o}-x_{i,o}||_{2}-||x_{j,c}-x_{i,c}||_{2}\},
    \label{eq:iso-rigidity}
\end{equation}
after adding the above objectives, our objective is :
\begin{equation}
    \mathcal{L}=\mathcal{L}_{image}+\lambda_{rigid}\mathcal{L}_{rigid}+\lambda_{rot}\mathcal{L}_{rot}+\lambda_{iso}\mathcal{L}_{iso}.
    \label{eq:losses}
\end{equation}
\subsection{Refinement Strategy}
\label{subsec:Optimization}
\textbf{Split-with-scale.}
Because the monocular video input for 3D-GS \cite{kerbl20233d} lacks multi-view supervision, a portion of the reconstructed point cloud (3D Gaussians) may become excessively sparse, leading to oversized Gaussians or blurring artifacts. To address this, we propose a strategy to split large Gaussians using a scale threshold $\epsilon_{scale}$. If a Gaussian has a size $s$ larger than $\epsilon_{scale}$, we decompose it into two identical Gaussians, each with half the size.

\noindent \textbf{Pose refinement.}
Despite the robust performance of 3D-GS \cite{kerbl20233d} in the presence of inaccurate SMPL parameters, there is a risk of generating a high-fidelity point cloud with inaccuracies. The inaccurate SMPL parameters may impact the model's alignment with the images, leading to blurred textures. To address this issue, we propose an optimization approach for the SMPL parameters. Specifically, we designate the SMPL pose parameters $\theta$ as the optimized parameter and refine them through the optimization process, guided by the defined losses.

\begin{table*}[!t]
\centering
\resizebox{\textwidth}{!}{
    \begin{tabular}{p{0.15\textwidth}cccccccccccc}
        \toprule
                    &\multicolumn{3}{c}{male-3-casual} & \multicolumn{3}{c}{male-4-casual} & \multicolumn{3}{c}{female-3-casual} & \multicolumn{3}{c}{female-4-casual}\\
                    &\textbf{PSNR↑} & \textbf{SSIM↑} & \textbf{LPIPS↓}  & \textbf{PSNR↑} & \textbf{SSIM↑} & \textbf{LPIPS↓} & \textbf{PSNR↑} & \textbf{SSIM↑} & \textbf{LPIPS↓} & \textbf{PSNR↑} & \textbf{SSIM↑} & \textbf{LPIPS↓} \\
        \midrule
             3D-GS\cite{kerbl20233d}          & 26.60 &0.9393 &0.082 &24.54 & 0.9469 & 0.088 &24.73 &0.9297 & 0.093 &25.74 &0.9364 &0.075 \\
             NeuralBody\cite{peng2021neural}   & 24.94 &0.9428 &0.033 &24.71 &0.9469 & 0.042 &23.87 &0.9504 & 0.035 &24.37 &0.9451 &0.038 \\
             Anim-NeRF\cite{chen2021animatable}      & 29.37 &0.9703 &0.017 &28.37 &0.9605 & \textbf{0.027} &28.91 &\textbf{0.9743}& \textbf{0.022} &28.90 &0.9678 &\textbf{0.017} \\
             InstantAvatar\cite{jiang2023instantavatar}  & 29.64 &0.9719 &\textbf{0.019} &28.03 &0.9647 &0.038 &28.27&0.9723& 0.025&29.58 &\textbf{0.9713} &0.020 \\
             Ours           & \textbf{35.66}&\textbf{0.9753}&0.021&\textbf{32.65} &\textbf{0.9769} & 0.049 &\textbf{33.22}&0.9701& 0.037&\textbf{31.43} &0.9630 &0.040 \\
        \bottomrule
    \end{tabular}}
    \caption{\textbf{Quantitative comparison of novel view synthesis on PeopleSnapshot dataset.} 
    Our approach exhibits a significant advantage in metric comparisons, showing substantial improvements in PSNR and SSIM metrics due to its superior restoration of image details.}
    \label{table:compare}
\end{table*}
\begin{figure*}
        \centering
        \includegraphics[width=1\linewidth]{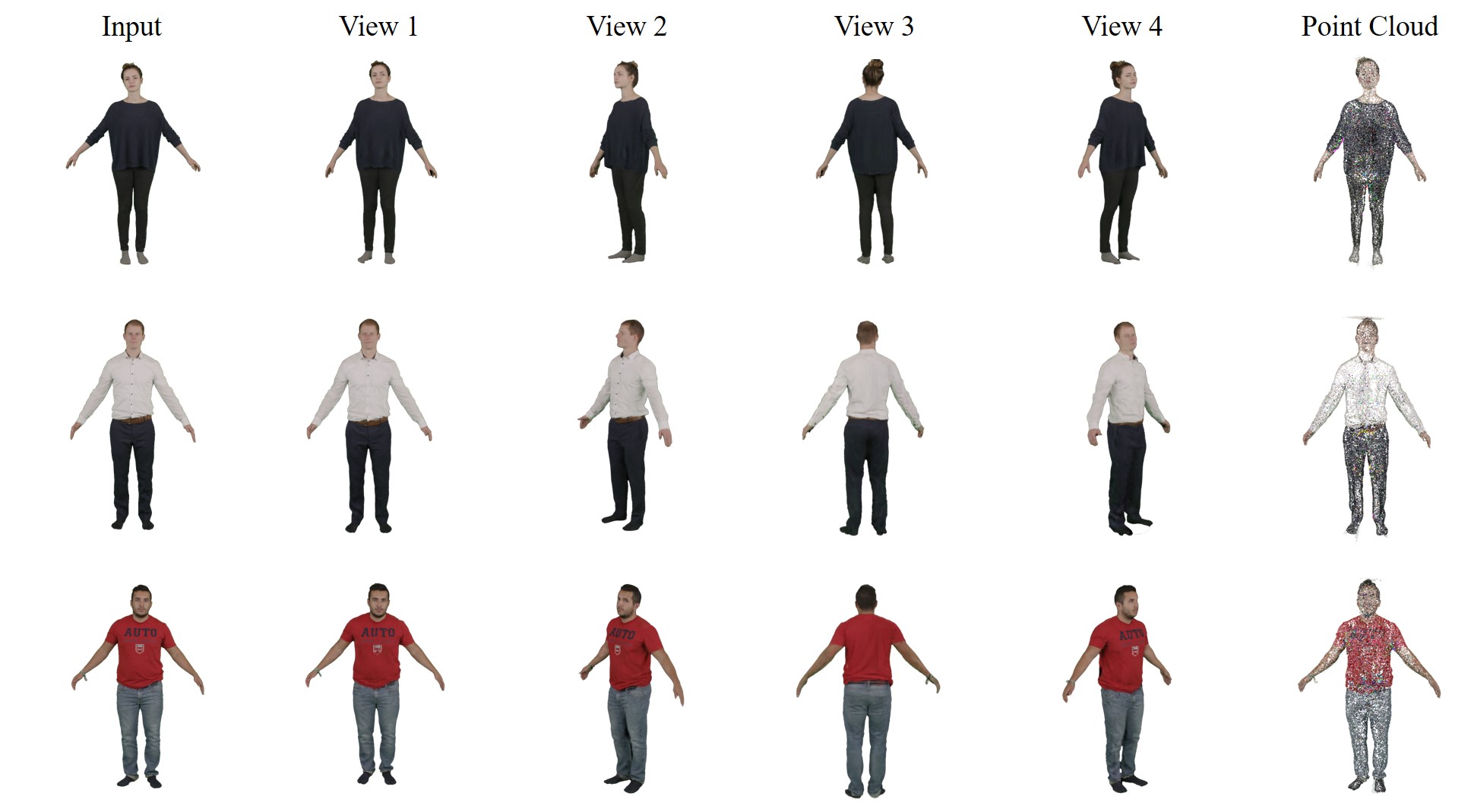}
        \caption{\textbf{Results of novel view synthesis and point cloud on PeopleSnapshot \cite{alldieck2018video} dataset.} Our method effectively restores details on the human body, including intricate details in the hair and folds on the clothes. Moreover, the generated point cloud faithfully captures geometric details on the clothing, demonstrating a commendable separation between geometry and texture.}
        \label{fig:novel-view}
\end{figure*}
\begin{figure*}
    \centering
    \includegraphics[width=1\linewidth]{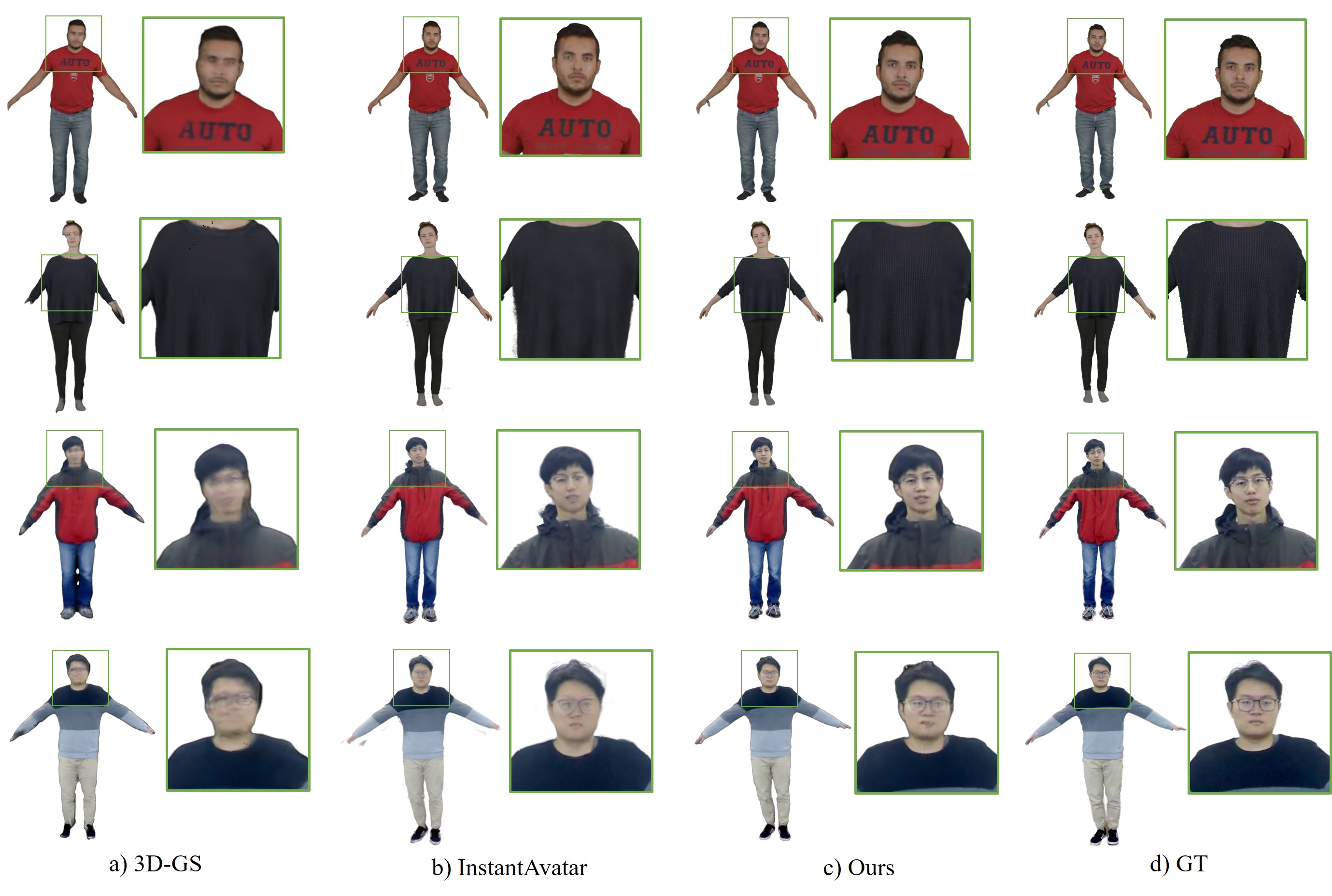}
    \caption{\textbf{Visual comparison} of different methods about novel view synthesis on PeopleSnapshot\cite{alldieck2018video}(column 1\&2) and iPER\cite{liu2019liquid}(column 3\&4). 
    3D-GS\cite{kerbl20233d} rely on multi-view consistency to gain center subjection which failed to handle dynamic scenes. 
    InstantAvatar\cite{jiang2023instantavatar} trade the quality and robust for time, might gain blur result on inaccurate parameters. 
    Our method gains high-fidelity results, especially on the cloth texture and the robustness.}
    \label{fig:visualcomapre}
\end{figure*}
\section{Experiment}
\label{sec:exp}
In this section, we evaluate our method on monocular training videos and compare it with the other baselines and state-of-the-art works. We also conduct ablation studies to verify the effectiveness of each component in our method.

\subsection{Datasets and Baseline}
\noindent \textbf{PeopleSnapshot.} PeopleSnapshot \cite{alldieck2018video} dataset contains eight sequences of dynamic humans wearing different outfits.
The actors rotate in front of a fixed camera, maintaining an A-pose during the recording.
We train the model with the frames of the human rotating in the first two circles and test with the resting frames.
The dataset also provides inaccurate shape and pose parameters. 
So we first process both the train dataset and test dataset to get the accurate pose parameters. 
Note that the coefficients of the Gaussians remain fixed. 
We evaluate the novel view synthesis quality with frame size in $540\times 540$ with the quantitative metrics including peak signal-to-noise ratio (PSNR), structural similarity index (SSIM), and learned perceptual image patch similarity (LPIPS), and train our model and other baselines in $1080\times 1080$ for visual comparison. 
The PeopleSnapshot dataset doesn’t have the corresponding ground truth point cloud, we only provide qualitative results.

\noindent \textbf{iPER.} iPER \cite{liu2019liquid} dataset consists of two sets of monocular RGB videos depicting humans rotating in an A-pose or engaging in random motions before a static camera. 
In the experiment, we leverage the A-pose series, adopting a similar setting to PeopleSnapshot. 
Subsequently, we visualize novel views and compare them with baselines to demonstrate the robustness of our method.

\noindent \textbf{Baselines.} We compare our method with original 3D-GS \cite{kerbl20233d}, Neural body\cite{peng2021neural}, Anim-NeRF \cite{chen2021animatable} and InstantAvatar \cite{jiang2023instantavatar}. 
Neural body\cite{peng2021neural} adopts the SMPL vertexes as the set of latent code to record the local feature and reconstruct humans in NeRF. 
Anim-nerf \cite{chen2021animatable} uses the explicit pose-guidance deformation that deforms the query point in observation space to canonical space with inverse linear blend skinning. 
InstantAvatar \cite{jiang2023instantavatar} builds the hash grid to restore the feature of NeRF and control the query points with Fast-SNARF \cite{chen2023fast}, which uses the root-finding way to find the corresponding points and transform it to the observation space to optimize the articulate weighting.
\subsection{Implementation Details}
GaussianBody is implemented in PyTorch and optimized with the Adam \cite{2014Adam}. 
We optimize the full model in 30k steps following the learning rate setting of official implementation, while the learning rate of position is initial in $6e^{-6}$ and the learning rate of pose parameters is $1e^{-3}$. 
We set the hyper-parameters as $\lambda_{rigid}=4e^{-2}$, $\lambda_{rot}=4e^{-2}$, $\lambda_{iso}=4e^{-2}$.
For training the model, it takes about 1 hour on a single RTX 4090.
\subsection{Results}
\textbf{Novel view synthesis. }
In Table \ref{table:compare}, our method consistently outperforms other approaches in various metrics, highlighting its superior performance in capturing detailed reconstructions. This indicates that our method excels in reconstructing intricate cloth textures and human body details.

Figure \ref{fig:visualcomapre} visually compares the results of our method with others. 3D-GS \cite{kerbl20233d} struggles with dynamic scenes due to violations of multi-view consistency, resulting in partial and blurred reconstructions. Our method surpasses InstantAvatar in cloth texture details, such as sweater knit patterns and facial features. Even with inaccurate pose parameters on iPER \cite{liu2019liquid}, our method demonstrates robust results. InstantAvatar's results, on the other hand, are less satisfactory, with inaccuracies in pose parameters leading to deformation artifacts.

Figure \ref{fig:novel-view} showcases realistic rendering results from different views, featuring individuals with diverse clothing and hairstyles. These results underscore the applicability and robustness of our method in real-world scenarios.

\paragraph{3D reconstruction.}
In Figure \ref{fig:novel-view}, the qualitative results of our method are visually compelling. The generated point clouds exhibit sufficient details to accurately represent the human body and clothing. Examples include the organic wrinkles in the shirt, intricate facial details, and well-defined palms with distinctly separated fingers. This level of detail in the point cloud provides a solid foundation for handling non-rigid deformations of the human body more accurately.
\subsection{Ablation Study}
\paragraph{Physically-based prior.}
To evaluate the impact of the physically-based prior, we conducted experiments by training models with and without the inclusion of both part-specific and holistic physically-based priors. Additionally, we visualized the model in the canonical space with the specified configurations.

\begin{figure}
    \centering
    \includegraphics[width=1\linewidth]{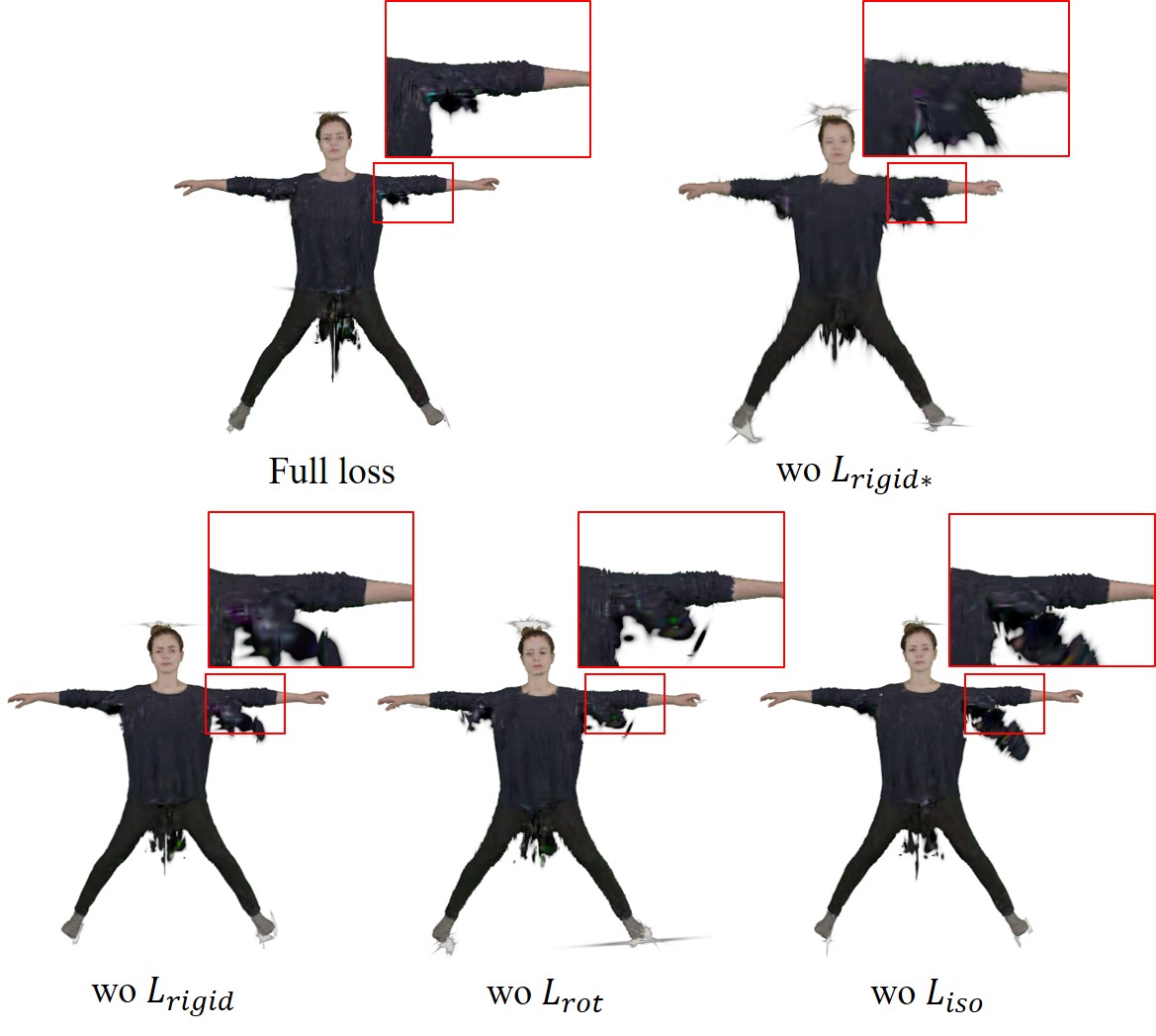}
    \caption{\textbf{Effect of physically-based prior. } This figure demonstrates what would be influenced without different losses. \
    Each loss influences different parts of the canonical model, the rigid loss regularizes a part of the wired rotation, the rot loss mainly reduces the glitch, and the iso loss reduces the unexpected transformation.}
    \label{fig:rigid-loss}
\end{figure}

Figure \ref{fig:rigid-loss} illustrates the results. In the absence of the physically-based prior, the model tends to produce numerous glitches, especially leading to blurred facial features. Specifically, the exclusion of the rigid loss contributes to facial blurring. On the other hand, without the rotational loss, the model generates fewer glitches, although artifacts may still be present. The absence of the isometric loss introduces artifacts stemming from unexpected transformations.

Only when incorporating all components of the physically-based prior, the appearance details are faithfully reconstructed without significant artifacts or blurring.

\begin{figure}
    \centering
    \includegraphics[width=1\linewidth]{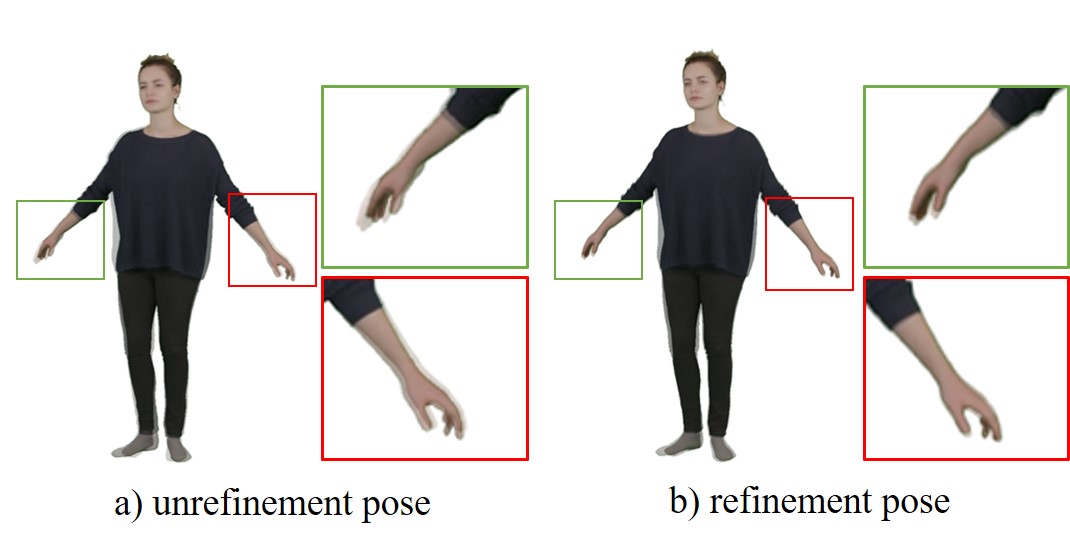}
    \caption{\textbf{Effect of pose refinement. } In the figure, the semi-transparent is the human appearance of our method. The hand part can demonstrate our method could adjust the pose.}
    \label{fig:refine}
\end{figure}

\begin{figure}
    \centering
    \includegraphics[width=1\linewidth]{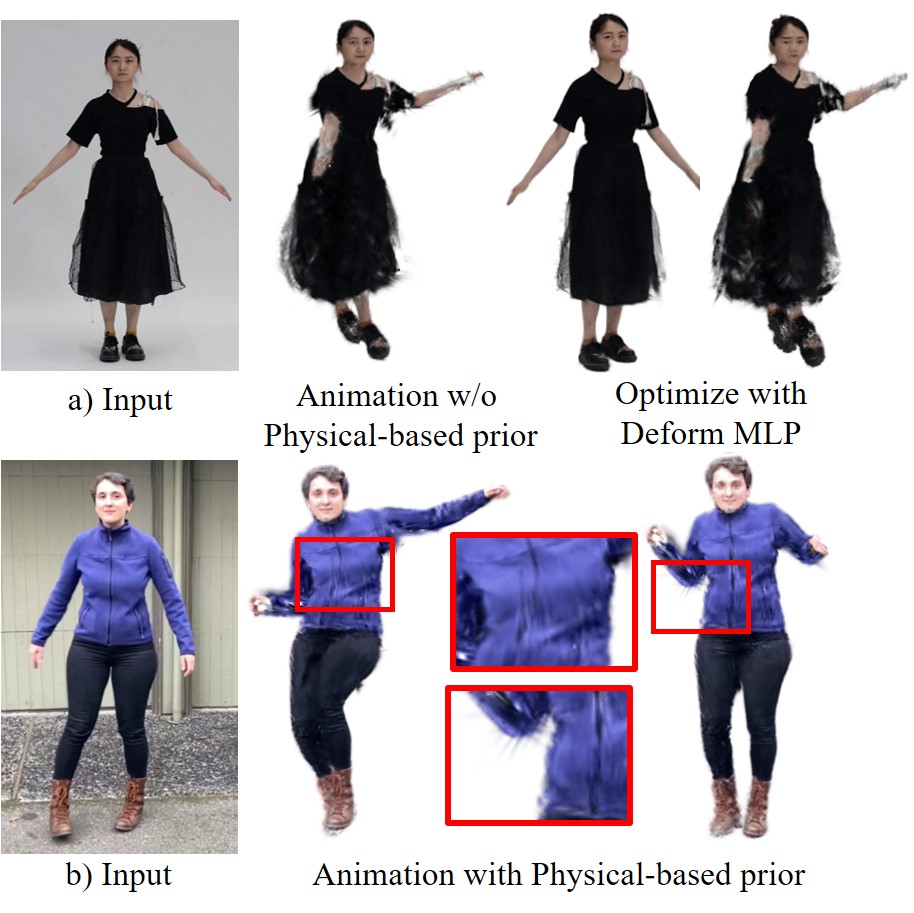}
    \caption{\textbf{Failure cases in deformation. } This picture demonstrates some results of our tries in deformation. 
    Due to conflicts between the deformation MLP and differentiable rendering, the model exhibits inaccuracies in both the canonical model and novel pose synthesis. 
    The Gaussians represent an elliptical region, leading to artifacts in the absence of precise deformation.
    }
    \label{fig:failcase}
\end{figure}

\paragraph{Pose refinement.}
We utilize SMPL parameters for explicit deformation, but inaccurate SMPL estimation can lead to incorrect Gaussian parameters, resulting in blurred textures and artifacts. Therefore, we introduce pose refinement, aiming to generate more accurate pose parameters, as depicted in Figure \ref{fig:refine}. This refinement helps mitigate issues related to blurred textures caused by misalignment in the observation space and avoids the need for the deformation MLP to fine-tune in each frame, as illustrated in Figure \ref{fig:failcase}.

\begin{figure}
        \centering
        \includegraphics[width=1\linewidth]{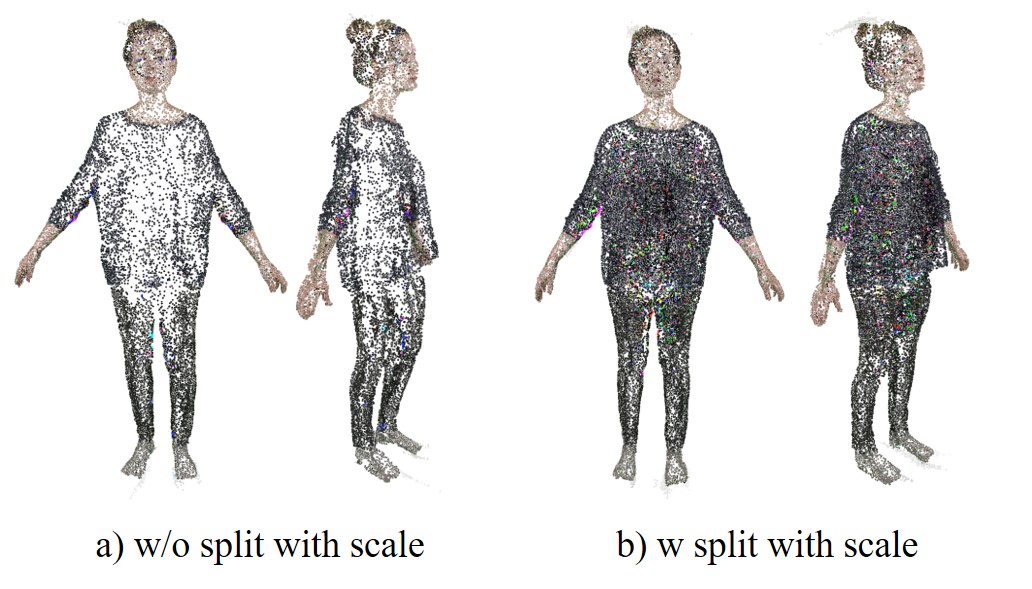}
        \caption{\textbf{Effect of splitting with the scale. } Without our optimization, the point cloud would be sparse. After the optimization, the point cloud could express more detail of the model}
        \label{fig:split}
\end{figure}

\paragraph{Split-with-scale.}
Given the divergence in our input compared to the original Gaussian input, especially the absence of part perspectives, the optimization process tends to yield a relatively sparse point cloud. This sparsity affects the representation of certain details during pose changes. As shown in Figure \ref{fig:split}, we address this issue by enhancing point cloud density through a scaling-based splitting approach.

\subsection{Discussion on Gaussian Deformation}
\paragraph{Deformation MLP.} 
We observe that the parameters after the deformation MLP network might become random, potentially misleading the optimization in the canonical space with differentiable rendering. To capture non-rigid deformation after the pose-guidance deformation, we adopt an approach inspired by SCARF \cite{feng2022capturing} and Deformable3dgs \cite{yang2023deformable3dgs} by introducing a deformation MLP network into the 3D-GS \cite{kerbl20233d} pipeline. We concatenate the Gaussians' point positions after the forward transform and their corresponding vertices into a fully connected neural network, aiming to obtain more accurate Gaussian parameters that capture the non-rigid deformation of the cloth in the observation space.

However, we encounter challenges where the canonical Gaussians lose generalization for novel views and pose synthesis, as illustrated in Figure \ref{fig:failcase}. The deformation MLP tends to overfit the observation space and even influences the result of the rigid transformation. This issue needs further investigation and optimization to achieve a more balanced representation.

\paragraph{Novel pose synthesis.}
The explicit representation offers several advantages, including accelerated training, simplified interfacing, and effective deformation handling. However, challenges arise from the sparse Gaussians and imprecise deformation, affecting the representation of novel poses. The sparsity of Gaussians, combined with the absence of accurate deformation, makes it challenging to represent a continuous surface. Despite attempts to mitigate these issues by reducing the size of Gaussians and applying regularization through the physically-based prior, unexpected glitches persist in novel poses, as illustrated in Figure \ref{fig:failcase}. Overcoming these challenges to reconstruct a reasonable non-rigid transformation of the cloth surface remains an area for improvement.
\section{Conclusion}
\label{sec:Conclusion}
In this paper, we present a novel method called GaussianBody for reconstructing dynamic clothed human models from monocular videos using the 3D Gaussians Splatting representation. By incorporating explicit pose deformation guidance, we extend the 3D-GS representation to clothed human reconstruction. To mitigate over-rotation issues between the observation space and the canonical space, we employ a physically-based prior to regularize the canonical space Gaussians. Additionally, we incorporate pose refinement and a split-with-scale strategy to enhance both the quality and robustness of the reconstruction. Our method achieves comparable image quality metrics with the baseline and other methods, demonstrating competitive performance, relatively fast training speeds, and the capability to train with higher resolution images.

{
    \small
    \bibliographystyle{unsrt}
    \bibliography{main}
}


\end{document}